\documentclass[runningheads]{llncs}
\usepackage[T1]{fontenc}
\usepackage{graphicx,verbatim}
\usepackage{amsfonts}

\begin{document}
\title{MuellerPT: Decomposition Driven Pre-training for Dense Learning in Mueller Polarimetry}
\titlerunning{MuellerPT: Decomposition Driven Pre-training in Mueller Polarimetry}
% \begin{comment}  %% Removed for anonymized MICCAI submission
\author{Adam Tlemsani\inst{1,2,3} \and
Yingdian Li\inst{2,3,4,5} \and
Maxime Giot\inst{2,3} \and Naim Slim\inst{2,3} \and Christopher J. Peters\inst{3} \and Abhijeet Ghosh\inst{1} \and Daniel S. Elson\inst{2,3}}
\authorrunning{A. Tlemsani et al.}
\institute{Department of Computing, Imperial College London \and Hamlyn Centre for Robotic Surgery, Imperial College London \and Department of Surgery and Cancer, Imperial College London \and Xi’an Institute of Optics and Precision Mechanics, Chinese Academy of Sciences \and University of Chinese Academy of Sciences  \\
\email{adam.tlemsani23@imperial.ac.uk
}}

% \end{comment}
  
\maketitle              % typeset the header of the contribution
\begin{abstract}
Mueller matrix imaging provides rich, physically meaningful contrast for biomedical tissue analysis, but supervised learning is hindered by scarce dense annotations and strong domain shifts across specimens and acquisition settings. We introduce MuellerPT, a physics guided pre-training approach that learns transferable dense representations by predicting Lu-Chipman decomposition maps from per-pixel 4x4 Mueller matrices. To scale pre-training, we collected a new large Multispectral Animal Polarimetric Organ dataset (MAP-Org). The pre-trained encoder is adapted with a segmentation head for grey vs. white matter segmentation in lamb brain. A classification head is used for colorectal cancer vs. non-cancer classification. Both segmentation and classification are evaluated across few-shot learning scenarios. In segmentation, MuellerPT improves label efficiency and cross specimen transfer compared to models without pre-training, achieving an absolute DICE gain of over 20\% compared to the baseline trained from scratch when using 5\% of the training data. In classification, MuellerPT also enhances label efficiency, improving overall accuracy by 8\% compared to the baseline when using 1\% of the training data. We demonstrate MuellerPT's robustness to domain shift with a qualitative evaluation of its predicted Lu-Chipman maps on an \textit{ex vivo} human oesophagus sample. These results suggest that predicting Lu-Chipman decomposition is an effective and practical pretext task for robust biomedical inference from Mueller polarimetry and can pave the way for future work on label efficient Mueller imaging. 

\keywords{Mueller Polarimetry  \and Pre-training \and Segmentation \and Classification}

\end{abstract}

\section{Introduction}

Polarization imaging encodes information that conventional intensity imaging is often insensitive to, particularly in turbid biological tissue where scattering, absorption anisotropy and tissue structural organisation affect the measured polarimetric response \cite{he2021polarisation}. In the Stokes-Mueller formalism, a $\textbf{M}\in\mathbb{R}^{4\times{4}}$ Mueller matrix describes the transformation of the incident Stokes vector into an emergent Stokes vector \cite{qi2017mueller}. Mueller Matrix Polarimetry (MMP) is particularly beneficial when microstructural changes in the tissue are diagnostic, such as the anisotropy changes evident in cancers including cervix, skin, colon, pancreas and brain \cite{he2021polarisation,robinson2023polarimetric,sampaio2023muller,ahmad2020mueller,hahne2025polarimetric}.

Despite the promise of MMP for a variety of medical diagnostic tasks, severe barriers impede the adoption of MMP systems including the time-sequential acquisition \cite{guyot2007registration}, large bulky systems \cite{chae2025machine} and how to efficiently utilise the Mueller matrix for clinical tasks \cite{he2021polarisation}. These constraints motivate the need for learning pipelines which are label efficient, robust to acquisition artefacts, resilient to domain shift and capable of producing representations which align with physical tissue contrast mechanisms.

The Lu-Chipman decomposition \cite{lu1996interpretation} factorises measured Mueller matrices into a sequence of matrices interpreted as a diattenuator, retarder and depolarizer, providing a clear route from 16 element Mueller matrices to a small set of physically interpretable parameters. Whilst including fewer parameters, Lu-Chipman decomposition can retain almost as much of the clinically relevant information as full 16 element MMP \cite{novikova2022complete}. This motivates working with these decompositions rather than the 16 element Mueller matrix as the parameters can be interpreted whilst maintaining much of the diagnostic information. 

These Lu-Chipman parameter maps are attractive for pre-training because they can be computed, without manual annotation, from unlabelled Mueller image acquisitions, providing structured supervision grounded in optics and a natural target which encourage encoders to learn features aligned with tissue microstructure. At the same time, these decomposition-derived maps are not error free but can be sensitive to measurement noise and depolarization behaviour, underscoring the need for robust learning objectives and evaluation \cite{huynh2021mueller}.

In polarimetric imaging, the data-driven polarimetry literature is expanding rapidly, but much of it remains supervised or focused on task-specific restoration rather than general purpose encoder pre-training \cite{yang2024data}. Explicit reusable pre-training schemes tailored to Mueller physics are scarce in the literature. Adjacent polarisation domains have demonstrated the utility of Self-Supervised Learning (SSL), for example, self-supervised contrastive learning has been proposed for 3D polarized light imaging of nerve fibre architecture \cite{oberstrass2024self}, yet to the best of our knowledge there exists no general Mueller pre-trained encoders.   

The core problem this paper addresses is that Mueller imaging models must learn informative, clinically transferable representations from high-dimensional polarimetric measurements under limited labels and diverse acquisition conditions (noise, calibration drift, device differences and tissue/domain shift) \cite{huynh2021mueller}. Existing medical SSL pipelines, while powerful, are typically agnostic to Mueller matrix information and many Mueller imaging studies either (1) train from scratch on small datasets (2) rely on generic model initialisations, or (3) use Lu-Chipman maps primarily as supervised features and not a pre-training signal.

We contribute the following.
\begin{enumerate}
    \item  MuellerPT, a novel pre-training pipeline for Mueller imaging encoders that uses Lu-Chipman parameter prediction as the pretext task 
    \item We demonstrate improved downstream segmentation and classification performance, particularly in the limited data setting, on representative Mueller imaging tasks compared to training from scratch with random weight initialisation 
    \item  We demonstrate robustness to domain shift as MuellerPT is resilient to various datasets and simulated systems
    \item A large open source dataset of reflected Mueller matrix images on a diverse range of animal tissue types set for release with this manuscript, named the Multispectral Animal Polarimetric Organ dataset (MAP-Org).
\end{enumerate}

\section{Methodology}

\begin{figure}
\includegraphics[width=\textwidth]{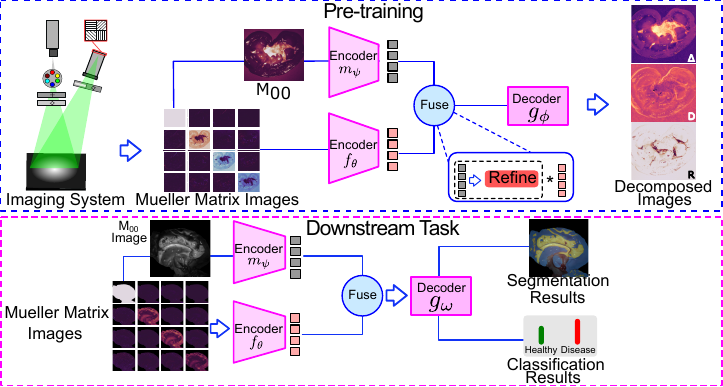}
\caption{MuellerPT outline - we first pre-trained on our new MAP-Org dataset and used Lu-Chipman prediction as our pretext task. The encoders for the Mueller and M00 images, $f_{\theta}$ and $m_{\psi}$ respectively, were then used and fine-tuned for the downstream tasks including segmentation and classification.} \label{MuellerPT}
\end{figure}

A dataset of 41 multispectral Mueller matrix images was acquired of fresh animal tissue (lamb heart/kidney/chops, chicken liver/breast/wings, steak, bacon) sourced from local butchers. The images were collected with a reflection configuration Mueller matrix imaging system. The system is broken down into two parts, a polarization state generator (PSG) and polarization state analyser (PSA).The PSG contains a collimated white light emitting diode (LED, MCWHLP3, Thorlabs, Newton, New Jersey, United States), a filter wheel with six different wavelengths ranging from 450 to 680 nm (FW102C, Thorlabs), a linear polarizer and a rotating quarter wave plate. This PSG generates four incident illumination polarization states. After the light reflects of the sample, a polarization sensitive camera (Blackfly S, model BFS-U3-51S5P-C, Teledyne FLIR) captures the reflected polarization state after passing through a slider containing either air or another quarter wave plate. Using a collection of known reference samples and the eigenvalue calibration method \cite{compain1999general}, the calibration error of our system was below 4.1\% for each wavelength. 

For a given measured Mueller image $\mathbf{M}\in\mathbb{R}^{H\times{W}\times{6}\times{4}\times{4}}$, where $H$ and $W$ denote the spatial dimensions, $6$ the number of wavelengths and $4\times{4}$ the elements of the Mueller matrix which will be denoted by $m(0,0)$ for the first element and $m(3,3)$ for the last. We first normalise our matrices by dividing out by $m(0,0)$ to get intensity independent representations. Measured Mueller matrices often lead to physically unrealisable matrices due to noise, saturation and poor alignment. We filter out these physically unrealisable matrices by testing for the condition that the eigenvalues of the coherency matrix are greater than or equal to 0 \cite{gil2016optimal}. Matrices which fail this test are projected to the closest physically-realisable matrix by clipping their negative eigenvalues to a small threshold of $1\times{10}^{-6}$.

\textbf{MuellerPT}. To enhance representation learning and downstream task performance we introduce MuellerPT, a novel model which utilises Lu-Chipman derived self-supervised learning.

Using MuellerPT is divided into two steps. We first pre-trained encoders $f_{\theta}$ and $m_{\psi}$ and decoder $g_{\phi}$ on our newly collected MAP-Org dataset. We then used the pre-trained encoders ($f_{\theta}$ and $m_{\psi}$) for a downstream task. For pre-training we exploited the popular Lu-Chipman decomposition to extract scalar values $\mathbf{\Delta}\in\mathbb{R}^{1}$, $\mathbf{R}\in\mathbb{R}^1$ and $\mathbf{D}\in\mathbb{R}^1$ from each Mueller matrix, representing the depolarization, retardance and diattenuation of the sample respectively. As these values can be computed numerically without manual labels, we used them as the pretext task to learn transferable representations.

To force greater representation learning and transferability, we applied Mueller element dropout during training. Instead of randomly selecting Mueller elements for dropout we removed elements which reflect physical hardware choices i.e using the upper left $\textbf{M}\in\mathbb{R}^{3\times3}$ subset of the Mueller matrix which reflects a system with no quarter-wave plates. Doing this allowed the model to learn representations which can transfer well to other tasks and systems.

We take inspiration from \cite{ma2021muellernet} and created two streams, the polarimetric stream ($f_{\theta}$) which encodes the normalised Mueller matrix and the normal stream ($m_{\psi}$) which encodes the $m(0,0)$ channel. This output vector is fed through a learnable refinement block before fusing with the polarimetric stream. The $m(0,0)$ element is unique in the Mueller matrix as it contains no polarization information instead encoding intensity information. Thus we choose to process them separately for two reasons. Firstly, each encoder can focus solely on either polarimetric or intensity images. Secondly, If a downstream task is using a dataset without the un-normalised $m(0,0)$, they can skip the normal stream and only use $f_{\theta}$.

To increase the size of our MAP-Org dataset when pre-training, we utilised augmentations to create additional Mueller images. Given the polarimetric nature of our images. We applied Mueller specific rotations \cite{hahne2025physically}  to our input images. To avoid heavy re-computation of the Lu-Chipman parameter ground truths after every rotation, we extended the work of \cite{hahne2025physically} to apply directly to the Lu-Chipman parameters instead of the Mueller matrix. This enabled us to skip having to recompute the Lu-Chipman parameters. The parameters of depolarization, retardance and diattenuation are invariant to viewing direction so we can apply simple spatial transformations to them. 

For a downstream task, we keep the encoders $f_{\theta}$ and $m_{\psi}$ and discard the decoder $g_{\phi}$. A new decoder, $g_{\omega}$, is initialised from scratch designed for the task at hand (i.e. segmentation or classification). During training for this new task, the pre-trained encoders were fine-tuned.

The MuellerPT pipeline is summarised in Figure \ref{MuellerPT}. For our network architecture we employed the HRNet \cite{wang2020deep} for its high resolution representations maps whose inductive bias match well with the pixel level task of Lu-Chipman prediction. To fairly compare our pre-trained encoders and decoders, we initialised an identical architecture from scratch with weights randomly initialised, and named this baseline HRNet-Scratch. Both HRNet-Scratch and MuellerPT followed the same training algorithm. Full implementation details will be provided in the code repository released alongside this manuscript.

\section{Results}

\subsubsection{Datasets:} For semantic segmentation evaluation we tested our model on the open source \textbf{PoLambRimetry} dataset \cite{10.1117/1.JBO.29.9.096002} which includes 20 multispectral images of different lamb brain regions from 6 different specimens. There are 16 labelled classes in the dataset although we follow the same processing as \cite{10.1117/1.JBO.29.9.096002} and grouped those into 2 classes: grey and white matter. There is an approximate 2:1 class imbalance in favour of grey matter. The PoLambRimetry dataset was chosen as it is, to the best of our knowledge, the only open source biomedical Mueller matrix dataset with pixel-level labels. For classification we evaluated on the \textbf{ColoPola} dataset \cite{pham2025colopola}. ColoPola contains 576 colorectal polarimetric slices with 288 healthy and 284 malignant samples. ColoPola was chosen for classification as it is, to the best of our knowledge, the largest open source biomedical Mueller matrix dataset with image level labels.

\textbf{Evaluation protocol:} For PoLambRimetry we evaluated every model on grey and white matter segmentation using specimen level splits and nested cross-validation. There were six specimens so we held one out for testing, another for validation and trained on the remaining four. This was repeated for every combination of train, validation and test sets. This nested cross validation approach ensured no images from the test or validation sets were used during training, preventing data leakage. As ColoPola doesn't have specimen level splits we divided the dataset into 60\% train, 20\% validation and 20\% test and computed metrics over 30 runs with different seeds.

\begin{figure}
\includegraphics[width=\textwidth]{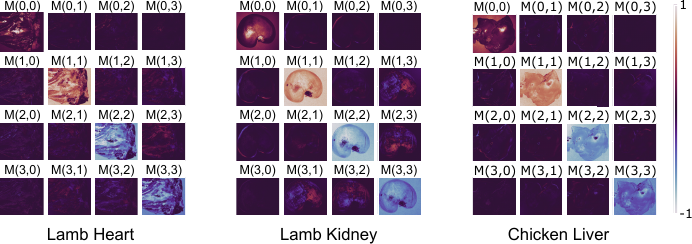}
\caption{Representative example Mueller matrices from our MAP-Org dataset. The m00 element displayed is un-normalized} \label{MAP-Org example}
\end{figure}

Representative example Mueller matrices from our MAP-Org dataset are presented in Figure \ref{MAP-Org example}.

We report both macro and per class DICE scores \cite{dice1945measures} for PoLambRimetry. We report overall accuracy, sensitivity and specificity for ColoPola. To assess performance in data limited scenarios, we report few shot performance for (1, 5, 25, 50 and 100 percent of the training data). 

The mean DICE scores for grey/white matter segmentation for both HRNet-Scratch and MuellerPT are presented in Table \ref{tab:polambrimetry_fewshot_dice_overall_gm_wm}, which shows that at the small data scenarios, MuellerPT achieves a much higher average DICE score compared to the HRNet-scratch with jumps of almost 6\% when using 1\% of the training data and 20\% when using 5\% of the training data. This suggests the Lu-Chipman prediction pretext task allows the pre-trained model to learn good representations which transfer well to the grey/white matter segmentation task. As the number of training samples increased the performance of the models are almost equal, with MuellerPT still having a slight edge.

\begin{table}
\caption{Few-shot grey/white matter segmentation on the PoLambRimetry dataset. Overall Dice, GM Dice, and WM Dice (mean $\pm$ std over $n=30$ runs).}
\label{tab:polambrimetry_fewshot_dice_overall_gm_wm}
\centering
\footnotesize
\begin{tabular}{lcccccc}
\hline
\textbf{Label(\%)} &
\multicolumn{3}{c}{\textbf{HRNet-Scratch}} &
\multicolumn{3}{c}{\textbf{MuellerPT (Ours)}} \\
 & \textbf{Overall} & \textbf{GM} & \textbf{WM} & \textbf{Overall} & \textbf{GM} & \textbf{WM} \\
\hline
1   & 0.47$\pm$0.09 & 0.71$\pm$0.15 & 0.23$\pm$0.18 & \textbf{0.53$\pm$0.11} & \textbf{0.74$\pm$0.12} & \textbf{0.32$\pm$0.20} \\
5   & 0.45$\pm$0.10 & 0.71$\pm$0.16 & 0.19$\pm$0.21 & \textbf{0.65$\pm$0.11} & \textbf{0.81$\pm$0.06} & \textbf{0.49$\pm$0.21} \\
25  & 0.77$\pm$0.07 & 0.87$\pm$0.05 & 0.67$\pm$0.12 & \textbf{0.78$\pm$0.07} & \textbf{0.88$\pm$0.05} & \textbf{0.68$\pm$0.12} \\
50  & 0.78$\pm$0.07 & 0.87$\pm$0.05 & 0.68$\pm$0.11 & \textbf{0.79$\pm$0.06} & \textbf{0.89$\pm$0.04} & \textbf{0.70$\pm$0.12} \\
100 & 0.80$\pm$0.06 & 0.89$\pm$0.04 & 0.70$\pm$0.12 & \textbf{0.80$\pm$0.06} & \textbf{0.90$\pm$0.03} & \textbf{0.70$\pm$0.12} \\
\hline
\end{tabular}
\end{table}

\begin{figure}
\includegraphics[width=\textwidth]{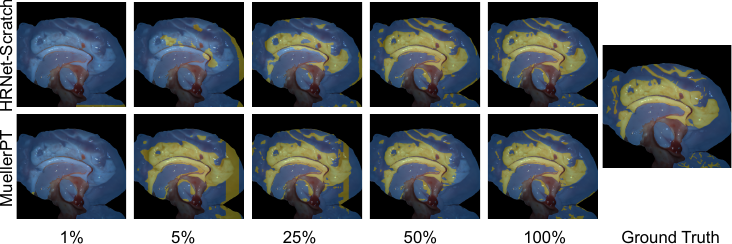}
\caption{Visualisations of the PoLambRimetry segmentation results for the HRNet-Scratch (top row) and our MuellerPT (bottom row) for different few shot learning scenarios. Grey matter is blue and white matter is yellow in this figure. The ground truth result is presented in the last column.} \label{vis}
\end{figure}

To understand how pre-training impacts both classes we report the per class DICE score in Table \ref{tab:polambrimetry_fewshot_dice_overall_gm_wm}, which shows that pre-training boosts the white matter performance the most at the small data setting with a jump of almost 30\% when using 5\% of the training data. This is encouraging as it shows pre-training can be especially beneficial to the minority class. As with the mean DICE score, as the number of training samples increased, the two methods approach each other in performance for the two classes, again with MuellerPT having the edge.

We provide a qualitative example of the segmented result of our MuellerPT model compared to the baseline for different N-shot learning scenarios in Figure \ref{vis}. Whilst both models struggle with this example at the extreme 1\% training budget, at 5\% our MuellerPT already demonstrates its ability to better predict the white matter in the central region of the brain. As the number of training examples increases, the two models converge and produce similar segmentation masks.

\begin{table}
\caption{Few-shot cancer vs.\ non-cancer classification on the ColoPola dataset. Overall Accuracy (Acc.), Sensitivity (Sens.; Cancer recall), and Specificity (Spec.; Non-cancer TNR) (mean $\pm$ std over $n=30$ runs).}
\label{tab:colopola_fewshot_acc_sens_spec}
\centering
\footnotesize
\renewcommand{\arraystretch}{1.15}
\begin{tabular}{lcccccc}
\hline
\textbf{Label (\%)} &
\multicolumn{3}{c}{\textbf{HRNet-Scratch}} &
\multicolumn{3}{c}{\textbf{MuellerPT (Ours)}} \\
 & \textbf{Acc.} & \textbf{Sens.} & \textbf{Spec.} & \textbf{Acc.} & \textbf{Sens.} & \textbf{Spec.} \\
\hline
1   & 0.51$\pm$0.04 & \textbf{0.84$\pm$0.34} & 0.18$\pm$0.34 & \textbf{0.59$\pm$0.09} & 0.64$\pm$0.34 & \textbf{0.54$\pm$0.37} \\
5   & 0.72$\pm$0.11 & \textbf{0.84$\pm$0.10} & 0.59$\pm$0.27 & \textbf{0.76$\pm$0.04} & 0.79$\pm$0.11 & \textbf{0.74$\pm$0.12} \\
25  & 0.84$\pm$0.03 & 0.86$\pm$0.05 & 0.82$\pm$0.05 & \textbf{0.85$\pm$0.04} & \textbf{0.86$\pm$0.05} & \textbf{0.83$\pm$0.08} \\
50  & 0.86$\pm$0.02 & 0.87$\pm$0.05 & 0.85$\pm$0.06 & \textbf{0.88$\pm$0.03} & \textbf{0.90$\pm$0.05} & \textbf{0.87$\pm$0.06} \\
100 & \textbf{0.90$\pm$0.03} & 0.90$\pm$0.04 & \textbf{0.91$\pm$0.05} & 0.90$\pm$0.03 & \textbf{0.92$\pm$0.04} & 0.88$\pm$0.06 \\
\hline
\end{tabular}
\end{table}

\begin{figure}
\includegraphics[width=\textwidth]{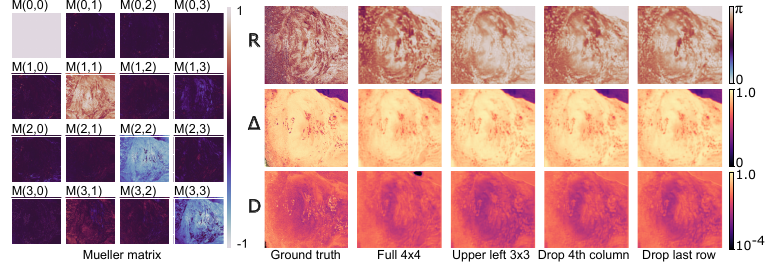}
\caption{Mueller matrix and MuellerPT Lu-Chipman decomposition Retardance (R), Depolarization ($\Delta$) and Diattenuation (D) results for a sample of oesophagus Mucinous adenocarcinoma (MAC) from a patient. The decompositions are computed for a set of reduced Mueller matrix configurations.} \label{hammersmith}
\end{figure}

We present the results for the ColoPola dataset in Table \ref{tab:colopola_fewshot_acc_sens_spec}. Pre-training helps the most in the low-label regime with jumps in overall accuracy of 8\% and 4\% when using 1\% and 5\% of training data respectively. HRNet-Scratch has very low specificity at 1\%, meaning it's likely predicting cancer too often (leading to a high false positive rate). At these low-label scenarios, the key benefit is a big jump in specificity (0.54 vs. 0.18 at 1\%) which raises the overall accuracy even though the sensitivity is lower.

To examine the transferability to human tissue, we explored the effectiveness of the pre-trained encoders ($f_{\theta}$, $m_{\psi}$) and decoder $g_{\phi}$ on a sample of oesophagus mucinous adenocarcinoma (MAC) collected \textit{ex vivo} from a patient at Hammersmith Hospital (Ethics number: 08/H0719/37) and imaged in the operating room. To assess the benefits of utilising the Lu-Chipman parameters as a pretext task, as opposed to simply using their numerical calculation, we evaluate the effectiveness of the pre-trained encoders and decoder in recovering these parameters in a variety of reduced Mueller matrix configurations, simulating realistic design choices of the imaging system such as no quarter-wave plates in the PSG/PSA. The results are shown in Figure \ref{hammersmith}. 

The pre-trained encoder and decoder are able to recover the Lu-Chipman parameters well, even with reduced configurations such as only using the upper left 3x3 of Mueller matrix. An upper left 3x3 configuration would not normally be able to numerically derive the Lu-Chipman parameters. This indicates our pre-trained models are able to generalise and transfer well, even to different simulated imaging systems and tissue domains not included in the pre-training.

\section{Conclusion}
This paper introduces MuellerPT, a novel method which leverages self-supervised learning for Mueller image segmentation and classification. We formulate Lu-Chipman parameter prediction as the pretext task. Experimental results demonstrate that MuellerPT outperforms an identical model trained from scratch, particularly in the small data setting with their performance being comparable at the larger data settings. This paves the way for more data and label-efficient models capable of capturing the best performance for the least amount of labels.

\begin{credits}
\subsubsection{\ackname} Adam Tlemsani is supported by UK Research and Innovation [UKRI AI Centre for Doctoral Training
in Digital Healthcare grant number EP/Y03097\\4/1]
Yingdian Li is supported by University of Chinese Academy of Sciences Joint PhD Training Program.
This work has benefited from infrastructure support from the National Institute for Health Research Imperial
Biomedical Research Centre, University of Chinese Academy of Sciences, the Wellcome Trust MedTechONE and
the Cancer Research UK Imperial Centre. The authors thank the clinical staff at Hammersmith Hospital for their assistance with specimen collection.
Ethics number 08/H0719/37

\subsubsection{\discintname}
The authors have no competing interests to declare that are
relevant to the content of this article.
\end{credits}

\bibliographystyle{splncs04}
\bibliography{mybibliography}

\end{document}